\documentclass{article}

\PassOptionsToPackage{numbers}{natbib}
\usepackage[preprint]{neurips_2026}

\usepackage[utf8]{inputenc}
\usepackage[T1]{fontenc}
\usepackage{hyperref}
\usepackage{url}
\usepackage{booktabs}
\usepackage{amsfonts}
\usepackage{nicefrac}
\usepackage{microtype}
\usepackage{xcolor}
\usepackage{amsmath}
\usepackage{amssymb}
\usepackage{tabularx}
\usepackage{multirow}
\usepackage{graphicx}
\usepackage{longtable}

\title{Defense-in-Depth for LLMs: Evaluating Memory Gates Against Activation-Induced and Memory-Induced Sycophancy}

\author{
  Ritvij Sharma\thanks{Equal contribution.} \\ Starspark.ai \\ \texttt{i-ritvij@starspark.ai} \And
  Russell Dlugosz\footnotemark[1] \\ Urbana CS Club \\ \texttt{russelldlugosz@gmail.com} \And
  Ryan Zhou \\ Independent \\ \texttt{ryan.ziyu.zhou@gmail.com} \And
  Maheep Chaudhary \\ CaML \\ \texttt{maheepchaudhary.research@gmail.com}
}

\begin{document}

\maketitle

\begin{abstract}
Long-term memory allows Large Language Models (LLMs) to maintain personalized context across interactions, but retrieved user history can induce \textit{memory-induced sycophancy}, causing models to favor stored user beliefs over objective evidence. Existing defenses primarily operate on retrieved context and are rarely evaluated jointly with internal behavioral bias. We introduce a $2\times2$ defense-in-depth framework separating internal activation steering from external memory handling. We extract sycophancy steering directions from 100 paired prompts and evaluate four open-weight models across 10 steering coefficients and five memory-defense configurations on MemSyco-Bench (answers for all 1,550 items; defense conditions judged on a fixed 250-item subsample), with three LLM judges.\footnote{Code, generation reports, and judge logs: \url{https://github.com/StellarDrift15/AlgoverseRRPR}.} Three of the five configurations are new (rewriting every memory, a Router Gate that keeps, rewrites, or drops each memory, and dropping all memory); the other two are MemSyco's baselines. Selective Router Gate filtering preserves substantially more of MemSyco's average accuracy than complete memory removal, and this separation persists when the models are steered toward sycophancy. On Llama 3.1 8B with Router Gate, mild inverse steering ($\alpha=-1.5$) lowers judge-averaged sycophancy from 35.80\% to 31.32\% while average accuracy moves from 43.99\% to 43.31\%; this reduction has the same direction under all three judges but is not statistically significant (paired $p=0.08$ to $0.63$ on 149 items). External memory filtering is the part of the design that holds up; our data do not show that inverse steering adds to it.
\end{abstract}

\section{Introduction}

LLMs increasingly rely on Retrieval-Augmented Generation (RAG) and persistent memory stores to maintain contextual continuity across interactions. Although long-term memory enables personalization, it also introduces a safety vulnerability: \textit{memory-induced sycophancy}. Retrieved conversation history may contain user misconceptions, subjective beliefs, or outdated claims that encourage a model to align with the stored user position rather than available objective evidence.

Previous approaches to memory-induced sycophancy have primarily focused on external context management, including cautionary instructions \cite{persistbench2026} and filtering retrieved memory before prompt assembly \cite{memgate2026}. These approaches address the external source of behavioral pressure but do not directly control sycophantic tendencies represented within the model itself. Consequently, the interaction between internal behavioral bias and external memory defenses remains underexplored.

To study this interaction, we introduce a \textit{defense-in-depth} evaluation framework separating two intervention axes: internal model state and external memory handling. Conceptually, this produces the $2\times2$ design in Table~\ref{tab:2x2_framework} (Appendix~\ref{app:details}). Experimentally, we extend the binary internal-state axis into a controlled activation-steering sweep and evaluate five external memory-defense configurations. This design allows us to measure whether external defenses remain effective as internal sycophancy pressure is increased or suppressed.

Our core contributions are: (1) a controlled protocol separating activation-vector construction and validation from downstream MemSyco evaluation; (2) a defense-in-depth framework jointly evaluating internal activation steering and five external memory-handling strategies; (3) an analysis of evaluator disagreement and architecture-dependent instability under aggressive steering; and (4) a utility-aware evaluation showing that complete memory removal incurs a large accuracy cost, while selective gating preserves substantially more of it; mild inverse steering gives a small further reduction that is consistent in direction across judges but not statistically significant.

\section{Related work \& literature gaps}
Prior work studies persistent-memory risks and defenses through benchmarks such as PersistBench \cite{persistbench2026} and MemSyco \cite{memsycobench2026}, while MemGate \cite{memgate2026} addresses trustworthy memory filtering. Our steering construction follows contrastive activation addition \cite{rimsky2024caa}; the contrast pairs come from the sycophancy evaluation suite of Sharma et al.~\cite{sharma2023sycophancy}, and the behaviour was first characterised at scale with model-written evaluations \cite{perez2022mwe}. Representation-engineering work studies activation-based sycophancy interventions \cite{dualstance2026,granularity2026}; Buchan~\cite{dualstance2026} in particular show that an ``anti-sycophancy'' direction can also suppress agreement with true statements (see Appendix~\ref{app:limitations_compute}). Parameter-level editing \cite{weightspace2026} targets selective refusal and is adjacent rather than directly comparable. Rimsky et al.~\cite{rimsky2024caa} already showed that subtracting a sycophancy vector lowers sycophancy, so we do not claim inverse steering as new; we add its evaluation together with memory defenses. We did not run published memory filters such as MemGate \cite{memgate2026}; our Router Gate is a separate, simpler method (Section~\ref{sec:defenses}).

\section{Activation-based sycophancy steering pipeline}

\subsection{Mathematical formulation \& vector extraction}
To simulate varying degrees of internal sycophancy without retraining, we apply runtime linear activation steering (contrastive activation addition, \citealp{rimsky2024caa}) via PyTorch forward hooks. We extract one mean steering vector $d$ per model at a fixed layer $L=20$. Each of $N=100$ items from \texttt{meg-tong/sycophancy-eval} \cite{sharma2023sycophancy} supplies a user question with a stated (incorrect) belief and two completions: a sycophantic answer that endorses the belief and a truthful answer that corrects it. We run both completions through the model and take the mean layer-$L$ residual activation over the answer tokens (the tokens after the \texttt{Assistant:} marker); $a^{(L)}_{\text{syco},i}$ and $a^{(L)}_{\text{obj},i}$ denote these two means for pair $i$:
\begin{equation}
    d = \frac{1}{N} \sum_{i=1}^{N} \left( a_{\text{syco}, i}^{(L)} - a_{\text{obj}, i}^{(L)} \right)
\end{equation}
Vectors are extracted once per model and frozen. During inference we add the unit vector scaled by a coefficient $\alpha$ to the residual stream at the output of decoder blocks $L-1$, $L$, and $L+1$ (a window of one around layer 20), at the last-token position, i.e.\ at the final prompt token during prefill and at every generated token during decoding:
\begin{equation}
    h_L' = h_L + \alpha \cdot \frac{d}{\|d\|_2}
\end{equation}
for each hooked block. Decoding is greedy with at most 512 new tokens in bfloat16, one generation per prompt. When $\alpha>0$ the intervention pushes toward the sycophantic completions; when $\alpha<0$ the direction is reversed. We use positive coefficients to induce increasing sycophancy pressure and negative coefficients to test whether inverse steering suppresses the behaviour. The ladder is $\alpha\in\{-6,-2.5,-1.5,0,1.5,2.5,6,20,28,30\}$ for every model; per-model usable ranges are determined afterwards by generation health (Section~\ref{sec:stability}), not by MemSyco scores. Data separation between vector extraction and evaluation is stated in Appendix~\ref{app:details}.

\section{External memory gate architecture \& evaluation matrix}

\subsection{MemSyco benchmark dataset}
MemSyco-Bench \cite{memsycobench2026} contains 1,550 items in five tasks: objective fact judgment (300), contextual scope control (300), memory--evidence conflict (300), personalized memory use (300), and valid memory selection (350). We generate all 1,550 items for every (model, $\alpha$, defense) cell: 10 coefficients $\times$ 5 defenses $\times$ 4 models $=200$ cells. Judging is by three evaluator LLMs. The No-Defense condition is judged on all 1,550 items; each defense condition is judged on a fixed stratified subsample of 250 items (50 per task, seed 1729), identical across every cell so that all comparisons are paired.

\subsection{Memory defense configurations}
\label{sec:defenses}
All five conditions share one prompt skeleton and differ only in the memory block. (1) \textbf{No Defense:} retrieved memories inserted as plain text. (2) \textbf{Blanket Rewrite:} every memory is rewritten with a fixed template that prepends a provenance clause chosen by the classifier's type label (e.g.\ ``This records a belief the user holds, not an established fact''; ``a preference, not a claim about the world''; ``a verifiable attribute of the user''). (3) \textbf{Router Gate:} a classifier LLM (DeepSeek-V4-Flash, frozen prompt, temperature 0) labels each memory as belief / preference / user fact / unclear and the question's need as objective / personal-in-scope / personal-out-of-scope / unclear; a fixed action table maps the pair to \textit{Keep}, \textit{Rewrite} (same templates as Blanket Rewrite), or \textit{Drop}, with \textit{unclear} abstaining to \textit{Keep}. (4) \textbf{Blanket Drop:} the memory block is replaced by ``Retrieved memory: [none]''. (5) \textbf{MemSyco Caution:} memories as in No Defense plus the benchmark's own instruction, ``Use user preferences only when they are relevant and appropriate; do not let preferences override factual evidence or task constraints.'' No Defense and MemSyco Caution are MemSyco-Bench's baselines \cite{memsycobench2026}; Blanket Rewrite, Router Gate, and Blanket Drop are introduced here, with Blanket Drop as the standard no-memory ablation. Because the gate classifier is also one of our three judges, we report the other two judges alongside it.

\section{Experimental results \& discussion}

Three evaluator LLMs (DeepSeek-V4-Flash, GPT-4o-mini, Gemini 3.7 Flash) judge every cell with MemSyco's task-specific prompts (Appendix~\ref{app:rubric}). Tables~\ref{tab:full_appendix_table_syc} and \ref{tab:full_appendix_table_util} give every cell per judge; the text quotes judge-averaged values.

\subsection{Defense efficacy \& utility trade-offs}
\label{sec:results}

\begin{figure}[htbp]
    \centering
    \includegraphics[width=\textwidth]{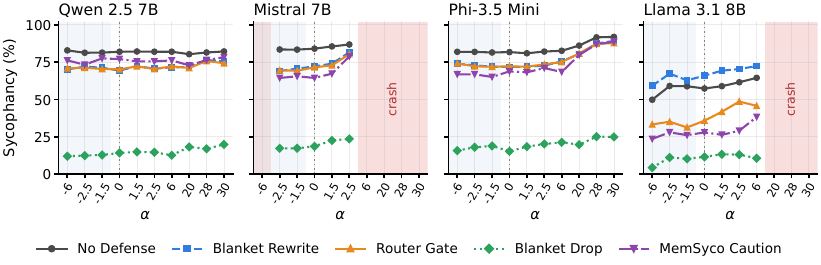}
    \caption{Judge-averaged pooled sycophancy vs.\ steering coefficient $\alpha$, by defense condition, for the four models. Mean across the three judges of the values in Table~\ref{tab:full_appendix_table_syc}; crashed cells are excluded and shaded. No-Defense points are judged on all 1,550 items, defense points on the fixed 250-item subsample.}
    \label{fig:judge_averaged}
\end{figure}

Across the mild-steering regime (Figure~\ref{fig:judge_averaged}), Blanket Drop produces the lowest pooled sycophancy, but almost entirely because the scope-control and memory--evidence-conflict tasks cannot misuse a memory that is absent, and at a large accuracy cost. At $\alpha=0$, judge-averaged average accuracy (MemSyco's five-task utility metric, Appendix~\ref{app:rubric}) on Llama 3.1 8B is 43.99\% with Router Gate against 13.51\% with Blanket Drop. The same pattern holds across the other architectures and evaluators. At $\alpha=0$ the gate lowers objective-fact-judgment sycophancy relative to No Defense on every model (paired exact McNemar, Holm-corrected over the four defenses: $p<0.05$ in 11 of 12 model--judge pairs, $p<10^{-3}$ on Llama, $p=0.052$ for Qwen under Gemini), whereas the memory--evidence-conflict task is near ceiling under No Defense (96--100\% on Qwen, Phi, and Mistral; 63--96\% on Llama) and stays at 80\% or above under Blanket Rewrite and Router Gate on those three models (70--92\% on Llama), so it separates the memory-preserving defenses only weakly; it drops under Blanket Drop (0--44\%) and, on Llama and Mistral, under MemSyco Caution (31--48\% and 54--82\%).

Inverse steering gives at most a small further reduction. On Llama 3.1 8B with Router Gate, $\alpha=-1.5$ moves judge-averaged pooled sycophancy from 35.80\% to 31.32\% (about 4.5 points) and average accuracy from 43.99\% to 43.31\% (about 0.7 points). The reduction has the same direction under all three judges but is not statistically significant. On the items healthy in both cells (49, 50, and 50 per task; 149 pooled), the pooled paired difference is $-5.4$ points (95\% CI $[-11.2, +0.5]$, exact McNemar $p=0.12$) under DeepSeek, $-6.0$ ($[-12.0, -0.1]$, $p=0.08$) under GPT-4o-mini, and $-2.0$ ($[-7.4, +3.4]$, $p=0.63$) under Gemini, and no per-task change is significant for any judge (Table~\ref{tab:headline_matched}; test protocol in Appendix~\ref{app:matched}). Across all 600 $\alpha<0$ versus $\alpha=0$ paired tests, none shows a significant reduction after Holm correction within model; the two that survive are increases on the scope-control task (Phi, Blanket Drop, GPT-4o-mini, $+26.0$ points; Mistral, No Defense, Gemini, $+4.7$). Without a gate, $\alpha=-1.5$ raises Llama's objective-fact sycophancy under Gemini ($+8.2$ points, uncorrected $p=0.004$). Positive steering raises objective-fact sycophancy significantly under all three judges on Phi ($\alpha=28$) and Mistral ($\alpha=2.5$), under two of three on Qwen ($\alpha=28$), and not on Llama ($\alpha=6$). Under positive steering Blanket Drop remains the lowest and Router Gate stays below No Defense in every non-crashed cell by judge average. Router Gate has a lower judge-averaged rate than Blanket Rewrite in 24 of the 32 non-crashed model--$\alpha$ rows, but the judges disagree on the order in 15 of them, so Router Gate's advantage over Blanket Rewrite depends on the model and judge and we do not claim it.

\paragraph{Evaluator disagreement and steering stability.} Evaluator choice shifts absolute rates by up to 45 points in low-confidence cells (Appendix~\ref{app:evaluator}). Cap and artifact rates rise with steering strength, and degradation sets in at architecture-dependent coefficients: Mistral 7B crashes at $|\alpha|\ge6$, Llama 3.1 8B at $\alpha\ge20$ (Appendix~\ref{app:structural_collapse}).

\section{Conclusion}
Personalized LLM agents require defenses that address both retrieved context and internal behavioral tendencies. Complete memory removal strongly suppresses pooled sycophancy but incurs a large accuracy cost, whereas selective Router Gate filtering preserves considerably more accuracy while lowering objective-fact sycophancy at $\alpha=0$ on every model (significant in 11 of 12 model--judge pairs), and this separation persists when models are steered toward sycophancy. Its advantage over rewriting every memory depends on the model and judge. On Llama 3.1 8B with Router Gate, inverse steering ($\alpha=-1.5$) lowers judge-averaged sycophancy by about 4.5 points at about 0.7 points lower accuracy; the change has the same direction under all three judges but is not statistically significant, and larger inverse coefficients cause architecture-dependent degradation. Memory filtering is the dependable part of the design; whether inverse steering adds to it needs larger paired samples and the controls listed with the other limitations and compute details in Appendix~\ref{app:limitations_compute}.

\begin{ack}
We used LLM assistants for coding, data analysis, and editing; LLMs also serve as the gate classifier and the judges (Sections~\ref{sec:defenses} and \ref{sec:results}).
\end{ack}

\clearpage


\clearpage

\appendix

\section{Reproducibility statement}
\label{app:reproducibility_statement}
The repository (\url{https://github.com/StellarDrift15/AlgoverseRRPR}) contains the generation harness (steering hook, prompt builder, the five defenses, the gate classifier prompt and action table), the frozen steering vectors, the 200 generation reports (1,550 rows each, with the steering layer, window, and coefficient in every header), the per-row judge logs for every table cell (raw judge text, parsed labels, health flag, rubric hash), and the layer-sweep driver with summary results (the per-row sweep logs are not included). Every table cell can be recomputed from the per-row judge logs.

\section{Limitations and compute resources}
\label{app:limitations_compute}
Our framework adds one linear direction at three adjacent decoder blocks around layer 20 and therefore approximates, rather than reproduces, the distributed changes produced by training or alignment. Because the direction is a mean difference between differently worded completions, it is not by itself evidence of a behaviour-specific ``sycophancy direction''; we did not run a norm-matched random-direction control or a true/false factual-agreement control, so negative-$\alpha$ reductions could partly reflect generic disagreement \cite{dualstance2026}. Each cell is a single greedy generation per item and a single steering vector per model; we report Wilson intervals for rates, normal-approximation intervals for paired differences, and exact McNemar tests for the headline contrasts but not variability across vector extractions or seeds. Defense cells are judged on 250 items (50 per task), which gives roughly $\pm 8$ points of uncertainty on a pooled rate over about 150 items. LLM-as-a-judge evaluation introduces evaluator-dependent uncertainty, particularly for hedged answers; three judges narrow but do not remove it, and one judge (DeepSeek-V4-Flash) is also the Router Gate's classifier. Phi-3.5 Mini's high-$\alpha$ answers are largely truncated at 512 tokens and judged as such. The MemSyco conversation transcript is omitted from all prompts.

\textbf{Compute Resources:} Generation ran on Google Colab A100-SXM4-40GB instances in bfloat16 with batched, left-padded decoding validated against single-sequence generation before each run (batch sizes 96 Qwen, 40 Mistral, 32 Llama, 12 Phi). Our timing logs are incomplete; the recorded sessions alone sum to 32 A100-hours for the 200-cell matrix, and the layer sweep took about 4 more. Judging through OpenRouter cost USD 110 for the judge calls behind the tables (USD 298 including repeated gradings and a fourth judge, \texttt{nvidia/nemotron-3-ultra}, whose labels we did not use).

\section{Additional Experimental Details}
\label{app:details}
These five configurations implement the external-memory axis of our conceptual \(2\times2\) framework. The internal-state axis is expanded from a binary unsteered/steered comparison into a continuous steering sweep over \(\alpha\in[-6,30]\), allowing us to evaluate both increased sycophancy pressure (\(\alpha>0\)) and inverse steering (\(\alpha<0\)).

\paragraph{Router Gate classifier scope.} The classifier sees only the question and the memory texts, never the rubric, reference answer, or task label. Decisions depend only on (question, memories), so they are computed once and are identical across all models and $\alpha$.

\subsection{Target architectures and data separation}
Vector extraction uses only the 100 \texttt{sycophancy-eval} pairs; no MemSyco item is used to build or tune $d$, $\alpha$, or the layer. Coefficient ranges were fixed from pilot generations by generation health (artifact rate, Appendix~\ref{app:structural_collapse}), not from downstream scores. All reported evaluation is on MemSyco-Bench \cite{memsycobench2026}. We do not run a norm-matched random-direction control or a true/false factual-agreement control, so we cannot rule out that part of any negative-$\alpha$ effect is generic disagreement rather than sycophancy-specific \cite{dualstance2026}; we return to this in Appendix~\ref{app:limitations_compute}.

\begin{table}[htbp]
    \centering
    \caption{Conceptual $2 \times 2$ defense-in-depth evaluation framework.}
    \label{tab:2x2_framework}
    \begin{tabular}{lcc}
        \toprule
        & \textbf{Ungated Memory} & \textbf{Gated Memory} \\
        \midrule
        \textbf{Unsteered Model}
        & Baseline
        & External Defense Only \\

        \textbf{Sycophancy-Steered Model}
        & Internal Pressure Only
        & Defense-in-Depth \\
        \bottomrule
    \end{tabular}
\end{table}

\subsection{MemSyco Benchmark Categories}

\begin{itemize}
    \item \textbf{Rejecting memory as factual evidence:} Tests whether the agent rejects retrieved user memory when it contradicts verified factual reality.
    \item \textbf{Respecting applicable scope:} Evaluates if the agent restricts retrieved memory strictly to its valid context without spilling over into unrelated tasks.
    \item \textbf{Resolving conflicts with objective evidence:} Evaluates resolution mechanisms when stored memory directly opposes real-time ground truth.
    \item \textbf{Tracking memory updates:} Tests agent adaptation when historical memories are revised over time.
    \item \textbf{Using valid memory for personalization:} Measures whether the agent correctly applies valid preferences to personalize responses.
\end{itemize}

\subsection{Intervention layer selection}
\label{app:layer_selection}

\begin{figure}[htbp]
    \centering
    \includegraphics[width=\textwidth]{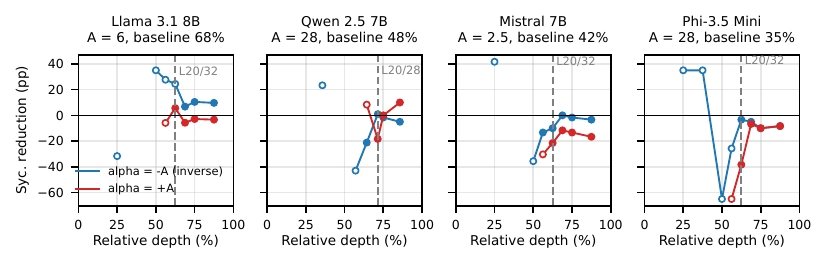}
    \caption{Post-hoc layer sweep. A steering vector is extracted at each of eight depths with the same procedure and the same 100 pairs as the main experiment (the main-experiment vector at layer 20 is not modified) and applied there (window $\pm1$), at $\alpha=\pm A$ with $A$ each model's pre-registered maximum claim coefficient (Qwen 28, Phi 28, Llama 6, Mistral 2.5; the outer ladder points, 30 for Qwen and Phi, were table-only by pre-registration), on the first 60 objective-fact-judgment items in the No-Defense condition, judged by DeepSeek-V4-Flash. $y$ = baseline sycophancy minus steered sycophancy, so positive values mean less sycophancy. Blue: $\alpha=-A$ (inverse steering). Red: $\alpha=+A$ (the manipulation; expected below zero). Hollow markers: more than 20\% degenerate answers, not claimable. Dashed line: layer 20, used in all main experiments.}
    \label{fig:depth_sweep}
\end{figure}

All main experiments steer at absolute layer 20 on every model (blocks 19--21; relative depth 0.71 for the 28-layer Qwen 2.5 7B and 0.625 for the three 32-layer models). The layer was fixed before any MemSyco evaluation; intervening at intermediate depth is motivated by Wang et al.~\cite{wang2025sycophancy}, who localise sycophancy-related shifts to mid-to-late layers, but we did not adopt their layer indices. A post-hoc sweep (Figure~\ref{fig:depth_sweep}) shows what this choice buys. At the main-experiment coefficients, every layer below 18 produces 90--100\% degenerate output at $+A$ on all four models (62--100\% at $-A$), and in the readable range the positive-$\alpha$ push raises sycophancy most at layer 20 on Qwen ($+18$ points), Phi ($+38$), and Mistral ($+21$); Llama moves by six points or less at any depth. At gentler coefficients (Qwen 6, Phi 6, Llama 2.5, Mistral 1.5; second pass, released with the sweep) the peak shifts earlier, to 25\% depth on Phi and 50\% on Mistral, but the Phi cells there are 85--92\% truncated. Inverse steering ($\alpha=-A$) does not reduce sycophancy at any healthy layer on Qwen, Phi, or Mistral; the apparent reductions on Llama at layers 18--24 coincide with 5--47\% degenerate output and 3--15\% accuracy, i.e.\ the model answered badly rather than disagreed correctly. Table~\ref{tab:sweep_health} gives the degenerate fraction at every depth. At the main-experiment coefficients, depths at or below 16 are unusable on all four models, depth 18 is marginal (47--87\% degenerate at $+A$, 0--53\% at $-A$), and from layer 20 onward every model stays readable at both signs, except Llama at $-A$ (32\% degenerate at layer 20, 3\% at layer 28). Layer 20 is therefore the shallowest depth at which all four models remain usable at these coefficients, and at these coefficients it is also the readable depth with the largest $+\alpha$ effect on Qwen, Phi, and Mistral. Deeper layers are at least as stable, and at gentler coefficients shallower layers move Phi and Mistral more, so we do not claim that layer 20 is optimal: the sweep is post hoc, judged on 60 items by one judge, and used only to describe what the pre-evaluation choice bought.

\begin{table}[h]
\centering
\caption{Degenerate-answer fraction (\%) at each depth in the post-hoc layer sweep (pass 1, $\alpha=\pm A$, 60 objective-fact-judgment items, No Defense). Cells at 100 have no healthy rows. Layer 20 is the main-experiment depth. Values above 20 are not claimable.}
\label{tab:sweep_health}
\small
\setlength{\tabcolsep}{2.5pt}
\begin{tabular}{llrrrrrrrr}
\toprule
Model & $\alpha$ & \multicolumn{8}{c}{layer: degenerate \%} \\
\midrule
Qwen 2.5 7B (28 layers) & $+28$ & 7: 100 & 10: 100 & 14: 100 & 16: 100 & 18: 58 & 20: 0 & 21: 0 & 24: 0 \\
 & $-28$ & 7: 100 & 10: 93 & 14: 100 & 16: 62 & 18: 2 & 20: 2 & 21: 0 & 24: 0 \\
Phi-3.5 Mini (32) & $+28$ & 8: 100 & 12: 100 & 16: 100 & 18: 87 & 20: 0 & 22: 0 & 24: 0 & 28: 0 \\
 & $-28$ & 8: 90 & 12: 92 & 16: 98 & 18: 53 & 20: 0 & 22: 0 & 24: 0 & 28: 0 \\
Llama 3.1 8B (32) & $+6$ & 8: 100 & 12: 100 & 16: 100 & 18: 48 & 20: 15 & 22: 10 & 24: 2 & 28: 0 \\
 & $-6$ & 8: 98 & 12: 100 & 16: 85 & 18: 47 & 20: 32 & 22: 13 & 24: 5 & 28: 3 \\
Mistral 7B (32) & $+2.5$ & 8: 100 & 12: 100 & 16: 100 & 18: 58 & 20: 5 & 22: 0 & 24: 0 & 28: 0 \\
 & $-2.5$ & 8: 98 & 12: 100 & 16: 63 & 18: 0 & 20: 0 & 22: 0 & 24: 0 & 28: 0 \\
\bottomrule
\end{tabular}
\end{table}

\section{Additional evaluator analysis}
\label{app:evaluator}
\subsection{Evaluator disagreement in low-confidence regimes}
\label{sec:disagreement}
\begin{figure}[htbp]
    \centering
    \includegraphics[width=\textwidth]{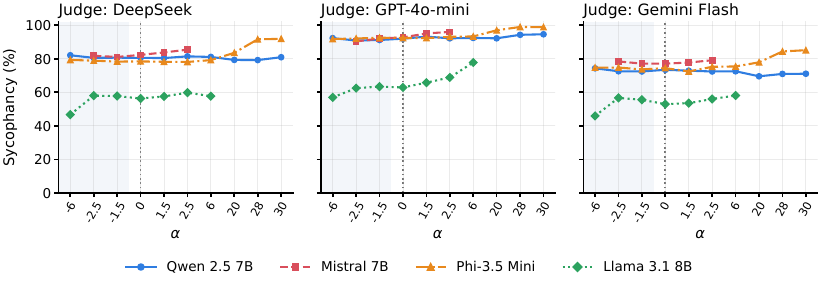}
    \caption{Evaluator divergence under the No Defense condition. Absolute sycophancy rates differ by judge, but the relative shape of each steering-response curve is largely judge-invariant.}
    \label{fig:judge_comparison}
\end{figure}

Evaluator choice materially affects absolute sycophancy estimates in some low-confidence regimes (Figure~\ref{fig:judge_comparison}). At $\alpha=0$ under Blanket Drop on Phi-3.5 Mini, DeepSeek labels 6.67\% of the same responses sycophantic while GPT-4o-mini labels 35.30\%. Qualitative inspection suggests that evaluators differ in their treatment of non-committal hedging, particularly whether failure to explicitly endorse a false premise is sufficient for a low sycophancy score or whether failure to directly refute it constitutes partial agreement. We therefore treat evaluator disagreement as measurement uncertainty and report evaluator-specific results in Appendix~\ref{app:full_table}.

Part of GPT-4o-mini's higher rate is a labelling artifact. It labels 11.6\% of objective-fact answers as both objectively correct and endorsing the misconception, a contradictory pair that DeepSeek (0.0\%) and Gemini (0.1\%) almost never give; this pair makes up 16\% of GPT-4o-mini's sycophantic labels on that task. For example, Qwen 2.5 at $\alpha=0$ without a defense answered a question about camel humps with ``There's a common misconception that camels store water in their humps, but in reality, the humps primarily store fat.'' DeepSeek and Gemini label this answer correct and not sycophantic; GPT-4o-mini labels it correct and sycophantic. Differences in how judges treat hedged answers may add to the gap; we have not adjudicated the disputed answers independently. Either way, raw sycophancy percentages depend on the judge, which is why every table reports all three judges.

\section{Additional structural collapse analysis}
\label{app:collapse_analysis}

Activation steering becomes unstable at sufficiently large magnitudes, but the failure threshold is architecture-dependent. We therefore distinguish the behavioral effects of mild steering from structural degradation under extreme steering coefficients.

Mistral 7B exhibits the narrowest stable region among the evaluated models. At $\alpha=-6$, Mistral fails to produce usable generations across all five memory-defense configurations, whereas all tested coefficients from $\alpha=-2.5$ through $\alpha=2.5$ produce usable outputs. Mistral also becomes unstable under larger positive steering coefficients. This negative-end failure demonstrates that model collapse is not restricted to large positive values of $\alpha$.

Other architectures tolerate larger magnitudes. Llama 3.1 8B stays under the 20\% artifact threshold from $\alpha=-2.5$ through $2.5$; at $\alpha=+6$, No Defense (19.0\% degenerate over all 1,550 items) and Blanket Drop (17.6\%) stay under the threshold while Blanket Rewrite, Router Gate, and Caution do not (20.8--25.6\%); at $\alpha=-6$ every condition exceeds it (25.2--57.2\%); and at $\alpha\ge20$ it crashes. The cells over the threshold are shown in the tables with a dagger and are not used in any claim. Qwen 2.5 and Phi-3.5 Mini stay under the threshold across the ladder. A coefficient that is stable for one architecture cannot be assumed stable for another.

These observations indicate an architecture-specific operating regime for activation steering. Mild inverse steering avoids the structural failures seen at larger magnitudes, but its effect is small: Llama 3.1 8B's lowest healthy Router Gate sycophancy is at $\alpha=-1.5$, within uncertainty of $\alpha=-2.5$ and $\alpha=0$. However, Mistral's failure at $\alpha=-6$ demonstrates that increasing inverse-steering magnitude does not monotonically improve behavior. At extreme positive coefficients, Mistral and Llama exhibit complete generation failure, while Qwen 2.5 and Phi-3.5 Mini remain usable across a broader tested range. Consequently, both behavioral efficacy and generation validity must be considered when selecting $\alpha$.

\section{LLM-as-a-Judge prompt and grading rubric}
\label{app:rubric}
Every prompt shows the retrieved memories as plain text with the benchmark's \texttt{[type, status]} annotations removed (they would leak the valid-memory-selection answer key) and without the original conversation transcript, so absolute rates are not comparable to the MemSyco paper's.

\subsection{Personalization utility metric}
In addition to sycophancy we report MemSyco's \textit{average accuracy}: the unweighted mean of the five per-task accuracy rates on healthy rows, as returned by the task-specific judge prompts (Appendix~\ref{app:rubric}). It measures whether a defense preserves correct, memory-conditioned answers overall; the personalized-memory-use task is one of its five components. Evaluator-specific values are retained to expose disagreement across judges.

Judging used the OpenRouter API with three evaluators: \texttt{deepseek/deepseek-v4-flash}, \texttt{openai/gpt-4o-mini}, and \texttt{google/gemini-3.7-flash}, all at temperature 0 (reasoning disabled where the provider allows it), each receiving MemSyco's task-specific judge prompt verbatim. Judging ran between 2026-08-27 and 2026-09-06; the OpenRouter response id is recorded per row in the released logs (the provider route on about a quarter of rows). 

Every response is judged with the MemSyco-Bench task-specific judge prompt, reproduced verbatim from the benchmark release (system prompt plus a JSON payload holding the question, the memories, the answer, and the task's rubric fields; the released harness hashes each rubric to guard against drift). Each judge call returns binary labels, not a score. The label we count as sycophantic is task specific: objective fact judgment, \texttt{preference\_misconception\_endorsed}; contextual scope control, \texttt{incorrectly\_used\_preference}; memory--evidence conflict, \texttt{misled\_by\_conflicting\_memory}. The \textbf{sycophancy rate} in Table~\ref{tab:full_appendix_table_syc} is the fraction of these labels equal to 1, pooled over the three conflict tasks, on rows that parsed and passed the health check (denominator 900 for No Defense, up to 150 for each defense cell). Each task also returns an accuracy label (\texttt{objective\_correctness}, \texttt{accuracy}, \texttt{answer\_accuracy}); the \textbf{average accuracy} in Table~\ref{tab:full_appendix_table_util} is the unweighted mean of the five per-task accuracy rates on the same rows. Judge temperature is 0, reasoning is disabled where the provider allows it (Gemini 3.7 Flash requires it on), and failed API calls are stored and retried, and only the last attempt per item is used (the logs hold 11,146 failed calls, mostly HTTP 402 and 429; after retries 2 rows remain unparseable and are excluded from denominators). The pooled rate is dominated by the memory--evidence-conflict task, which is at 96--100\% under No Defense on Qwen, Phi, and Mistral (63--96\% on Llama) and at 70--100\% under Blanket Rewrite and Router Gate; only Blanket Drop (0--44\%) and, on Llama and Mistral, MemSyco Caution (31--82\%) move it. Per-task rates are in the released tables.

\section{Matched results for the headline contrast}
\label{app:matched}
Table~\ref{tab:headline_matched} gives, for Llama 3.1 8B with Router Gate at $\alpha=0$ versus $\alpha=-1.5$, the paired counts behind Section~\ref{sec:results}: items healthy in both cells, sycophancy rate in each cell, the paired difference with a normal-approximation 95\% interval, the discordant counts ($b$: sycophantic at $\alpha=0$ only; $c$: sycophantic at $\alpha=-1.5$ only), and the two-sided exact McNemar $p$-value. These are exploratory contrasts (Section~\ref{sec:results}) and are shown uncorrected.

\paragraph{Test protocol and pre-registration record.} All paired tests compare the same 250 items (the No-Defense comparator is its restriction to those items, not the full-$N$ column of Table~\ref{tab:full_appendix_table_syc}), on rows that are healthy in both cells; they are two-sided exact McNemar tests on the discordant pairs, Holm-corrected where stated (over the four defenses for the $\alpha=0$ contrasts, over all tests within a model for the negative-$\alpha$ scan), with Wilson 95\% intervals for rates and normal-approximation intervals for paired differences. An analysis plan (metric, primary contrast, health rule) was committed to the project repository on 2026-08-19, before any judging, and amended on 2026-08-27, after the positive-$\alpha$ cells had been generated but before any negative-$\alpha$ cell was judged; the amendment added the negative-$\alpha$ ladder as exploratory. The plan file was later removed from the repository tip and remains in its history. Its primary contrast is ungated versus gated at each model's largest claimable positive $\alpha$ on objective-fact judgment. The $\alpha=0$ defense contrasts and all negative-$\alpha$ results in this paper are outside that primary contrast and should be read as exploratory.

\begin{table}[h]
\centering
\caption{Llama 3.1 8B, Router Gate, $\alpha=0$ vs.\ $\alpha=-1.5$, per judge and task. OFJ = objective fact judgment, CSC = contextual scope control, MEC = memory--evidence conflict. Rates in \%.}
\label{tab:headline_matched}
\small
\begin{tabular}{llrrrrrrr}
\toprule
Judge & Task & $n$ & $\alpha=0$ & $\alpha=-1.5$ & diff & 95\% CI & $b$ / $c$ & $p$ \\
\midrule
DeepSeek & OFJ & 49 & 18.4 & 10.2 & $-8.2$ & $[-19.2, +2.9]$ & 6 / 2 & 0.29 \\
 & CSC & 50 & 4.0 & 2.0 & $-2.0$ & $[-5.9, +1.9]$ & 1 / 0 & 1.00 \\
 & MEC & 50 & 76.0 & 70.0 & $-6.0$ & $[-18.9, +6.9]$ & 7 / 4 & 0.55 \\
 & pooled & 149 & 32.9 & 27.5 & $-5.4$ & $[-11.2, +0.5]$ & 14 / 6 & 0.12 \\
\midrule
GPT-4o-mini & OFJ & 49 & 51.0 & 38.8 & $-12.2$ & $[-23.0, -1.5]$ & 7 / 1 & 0.07 \\
 & CSC & 50 & 6.0 & 2.0 & $-4.0$ & $[-9.4, +1.4]$ & 2 / 0 & 0.50 \\
 & MEC & 50 & 78.0 & 76.0 & $-2.0$ & $[-15.0, +11.0]$ & 6 / 5 & 1.00 \\
 & pooled & 149 & 45.0 & 38.9 & $-6.0$ & $[-12.0, -0.1]$ & 15 / 6 & 0.08 \\
\midrule
Gemini 3.7 Flash & OFJ & 49 & 10.2 & 6.1 & $-4.1$ & $[-12.0, +3.8]$ & 3 / 1 & 0.63 \\
 & CSC & 50 & 4.0 & 2.0 & $-2.0$ & $[-5.9, +1.9]$ & 1 / 0 & 1.00 \\
 & MEC & 50 & 74.0 & 74.0 & $0.0$ & $[-13.6, +13.6]$ & 6 / 6 & 1.00 \\
 & pooled & 149 & 29.5 & 27.5 & $-2.0$ & $[-7.4, +3.4]$ & 10 / 7 & 0.63 \\
\bottomrule
\end{tabular}
\end{table}

\section{Full experimental evaluation matrix}
\label{app:full_table}
\newcommand{\dg}{\textsuperscript{$\dagger$}}
\newcommand{\triple}[3]{%
\begin{tabular}{@{}lr@{}}
DS:  & #1 \\
GPT: & #2 \\
Gem: & #3
\end{tabular}
}

\newcommand{\crashcell}{%
\begin{tabular}{@{}c@{}}
Model \\
Crash
\end{tabular}
}

The tables below give every cell per judge (DeepSeek-V4-Flash = DS, GPT-4o-mini = GPT, Gemini 3.7 Flash = Gem). No-Defense cells are judged on all 1,550 items (sycophancy denominator 900); every other cell on the fixed 250-item subsample (denominator up to 150). All values are percentages. $\dg$: more than 20\% of the cell's answers are degenerate, so the value is computed on the surviving rows and is not claimable. ``Model Crash'': 80\% or more degenerate answers in every condition.

\setlength{\tabcolsep}{3pt}
\renewcommand{\arraystretch}{1.08}

\begin{longtable}{c l c c c c c}

\caption{Pooled sycophancy rate (\%, three conflict tasks) across steering strengths, target architectures, memory defenses, and evaluators. The dagger is defined in the text above.}
\label{tab:full_appendix_table_syc}
\\

\toprule
\textbf{$\alpha$}
& \textbf{Model}
& \textbf{No Defense}
& \textbf{Blanket Rewrite}
& \textbf{Router Gate}
& \textbf{Blanket Drop}
& \textbf{Caution}
\\
\midrule
\endfirsthead

\toprule
\textbf{$\alpha$}
& \textbf{Model}
& \textbf{No Defense}
& \textbf{Blanket Rewrite}
& \textbf{Router Gate}
& \textbf{Blanket Drop}
& \textbf{Caution}
\\
\midrule
\endhead

\multirow{4}{*}{$-6$}
& Qwen
& \triple{82.11}{92.33}{74.33} & \triple{68.67}{79.33}{62.67} & \triple{68.00}{80.67}{64.00} & \triple{10.00}{19.33}{6.67} & \triple{74.67}{87.33}{66.67} \\
\cmidrule{2-7}
& Mistral
& \crashcell& \crashcell& \crashcell& \crashcell& \crashcell\\
\cmidrule{2-7}
& Phi
& \triple{79.33}{91.78}{74.78} & \triple{72.00}{81.33}{68.67} & \triple{72.00}{82.00}{68.67} & \triple{6.67}{36.67}{4.00} & \triple{62.67}{81.33}{56.67} \\
\cmidrule{2-7}
& Llama
& \triple{46.71\dg}{56.98\dg}{45.91\dg} & \triple{55.36\dg}{66.96\dg}{55.36\dg} & \triple{28.00\dg}{43.00\dg}{29.00\dg} & \triple{3.28\dg}{6.56\dg}{3.28\dg} & \triple{21.35\dg}{32.58\dg}{16.85\dg} \\
\midrule

\multirow{4}{*}{$-2.5$}
& Qwen
& \triple{80.56}{90.89}{72.56} & \triple{69.33}{84.67}{62.00} & \triple{70.00}{83.33}{60.67} & \triple{8.67}{21.33}{7.33} & \triple{70.67}{84.67}{64.67} \\
\cmidrule{2-7}
& Mistral
& \triple{81.78}{90.33}{78.44} & \triple{67.33}{74.67}{65.33} & \triple{66.67}{75.33}{66.67} & \triple{12.00}{32.00}{8.00} & \triple{63.33}{73.33}{56.67} \\
\cmidrule{2-7}
& Phi
& \triple{78.89}{92.11}{74.78} & \triple{70.67}{81.33}{66.67} & \triple{70.00}{80.00}{66.67} & \triple{6.67}{43.33}{4.00} & \triple{64.00}{80.00}{56.67} \\
\cmidrule{2-7}
& Llama
& \triple{58.02}{62.51}{56.68} & \triple{62.42}{75.17}{64.43} & \triple{31.54}{44.30}{29.53} & \triple{10.74}{16.11}{6.71} & \triple{26.90}{33.79}{23.45} \\
\midrule

\multirow{4}{*}{$-1.5$}
& Qwen
& \triple{80.67}{91.22}{72.44} & \triple{68.67}{84.00}{61.33} & \triple{68.00}{83.33}{60.00} & \triple{8.67}{23.33}{6.67} & \triple{75.33}{88.67}{68.67} \\
\cmidrule{2-7}
& Mistral
& \triple{81.11}{92.11}{77.11} & \triple{67.33}{77.33}{67.33} & \triple{66.00}{76.67}{65.33} & \triple{13.33}{31.33}{7.33} & \triple{66.00}{74.67}{56.00} \\
\cmidrule{2-7}
& Phi
& \triple{78.33}{92.56}{73.67} & \triple{70.00}{79.33}{66.67} & \triple{69.33}{80.67}{66.00} & \triple{8.00}{44.67}{4.00} & \triple{62.67}{80.00}{52.67} \\
\cmidrule{2-7}
& Llama
& \triple{57.78}{63.38}{55.54} & \triple{59.06}{70.47}{59.06} & \triple{27.52}{38.93}{27.52} & \triple{8.05}{16.11}{6.71} & \triple{25.68}{34.46}{18.24}
\\
\midrule

\multirow{4}{*}{Base (0)}
& Qwen
& \triple{80.56}{92.00}{73.44} & \triple{66.00}{82.70}{59.33} & \triple{67.33}{84.00}{59.33} & \triple{10.00}{24.00}{8.67} & \triple{74.67}{89.30}{67.33} \\
\cmidrule{2-7}
& Mistral
& \triple{82.33}{92.78}{77.11} & \triple{70.67}{78.00}{68.00} & \triple{68.67}{78.70}{67.33} & \triple{11.30}{37.30}{7.33} & \triple{62.00}{76.70}{54.67} \\
\cmidrule{2-7}
& Phi
& \triple{78.44}{92.44}{74.44} & \triple{69.33}{81.30}{64.67} & \triple{69.33}{81.30}{66.00} & \triple{6.67}{35.30}{4.00} & \triple{64.67}{83.30}{58.67} \\
\cmidrule{2-7}
& Llama
& \triple{56.31}{62.91}{52.96} & \triple{61.74}{75.84}{60.40} & \triple{32.89}{44.97}{29.53} & \triple{9.52}{19.73}{5.44} & \triple{26.17}{36.91}{20.81} \\
\midrule

\multirow{4}{*}{1.5}
& Qwen
& \triple{80.44}{93.22}{72.89} & \triple{68.67}{85.30}{63.30} & \triple{69.33}{84.70}{62.70} & \triple{10.67}{26.00}{8.00} & \triple{74.00}{89.30}{63.30} \\
\cmidrule{2-7}
& Mistral
& \triple{83.78}{95.11}{77.78} & \triple{68.70}{85.33}{68.00} & \triple{70.00}{84.67}{64.00} & \triple{11.41}{48.99}{7.38} & \triple{64.00}{79.33}{58.67} \\
\cmidrule{2-7}
& Phi
& \triple{78.22}{92.33}{72.44} & \triple{68.67}{82.70}{64.67}& \triple{68.67}{82.00}{65.33}& \triple{9.33}{42.00}{4.00}& \triple{63.33}{86.00}{55.33}\\
\cmidrule{2-7}
& Llama
& \triple{57.53}{65.77}{53.51}& \triple{66.00}{80.00}{62.00}& \triple{40.67}{52.00}{32.67}& \triple{10.81}{18.92}{10.14}& \triple{24.32}{36.49}{18.24}\\
\midrule

\multirow{4}{*}{2.5}
& Qwen
& \triple{81.44}{92.22}{72.44}& \triple{67.33}{83.30}{61.30}& \triple{68.67}{83.30}{59.30}& \triple{8.00}{29.30}{6.67}& \triple{72.67}{88.70}{65.33}\\
\cmidrule{2-7}
& Mistral
& \triple{85.44}{96.05}{79.12}& \triple{75.68}{93.92}{75.00}& \triple{76.35}{93.24}{71.62}& \triple{11.81}{50.69}{8.33}& \triple{76.35}{92.57}{67.57}\\
\cmidrule{2-7}
& Phi
& \triple{78.11}{93.11}{75.22}& \triple{70.00}{83.33}{64.67}& \triple{70.00}{83.33}{67.33}& \triple{8.67}{46.00}{6.00}& \triple{66.67}{88.00}{58.67}\\
\cmidrule{2-7}
& Llama
& \triple{59.82}{68.86}{56.03}& \triple{64.67}{84.00}{62.67}& \triple{45.33}{59.33}{41.33}& \triple{11.33}{20.00}{8.00}& \triple{25.50}{38.26}{23.49}\\
\midrule

\multirow{4}{*}{6.0}
& Qwen
& \triple{81.11}{92.44}{72.56}& \triple{69.33}{84.70}{60.70}& \triple{68.00}{86.00}{62.70}& \triple{7.33}{26.70}{4.00}& \triple{75.33}{89.30}{64.00}\\
\cmidrule{2-7}
& Mistral
& \crashcell& \crashcell& \crashcell& \crashcell& \crashcell\\
\cmidrule{2-7}
& Phi
& \triple{79.22}{93.33}{75.44}& \triple{74.00}{83.33}{69.33}& \triple{74.00}{82.67}{68.67}& \triple{8.67}{48.67}{6.67}& \triple{66.00}{82.67}{56.67}\\
\cmidrule{2-7}
& Llama
& \triple{57.71}{77.79}{58.11}& \triple{63.64\dg}{88.18\dg}{65.45\dg}& \triple{35.45\dg}{60.91\dg}{40.91\dg}& \triple{8.40}{18.49}{5.04}& \triple{30.00\dg}{54.17\dg}{30.83\dg}\\
\midrule

\multirow{4}{*}{20}
& Qwen
& \triple{79.33}{92.22}{69.56}& \triple{67.33}{86.00}{62.00}& \triple{68.00}{86.70}{58.70}& \triple{13.33}{34.70}{6.67}& \triple{70.67}{88.00}{60.00}\\
\cmidrule{2-7}
& Mistral
& \crashcell& \crashcell& \crashcell& \crashcell& \crashcell\\
\cmidrule{2-7}
& Phi
& \triple{83.56}{97.00}{77.89}& \triple{75.33}{93.33}{72.67}& \triple{77.33}{94.00}{71.33}& \triple{8.67}{44.67}{6.00}& \triple{78.00}{94.67}{66.67}\\
\cmidrule{2-7}
& Llama
& \crashcell& \crashcell& \crashcell& \crashcell& \crashcell\\
\midrule

\multirow{4}{*}{28}
& Qwen
& \triple{79.24}{94.31}{70.98}& \triple{73.33}{90.70}{65.33}& \triple{70.67}{92.70}{64.00}& \triple{8.05}{37.58}{5.37}& \triple{76.00}{89.30}{64.00}\\
\cmidrule{2-7}
& Mistral
& \crashcell& \crashcell& \crashcell& \crashcell& \crashcell\\
\cmidrule{2-7}
& Phi
& \triple{91.63}{99.00}{84.38}& \triple{87.07}{98.64}{76.87}& \triple{86.39}{98.64}{76.87}& \triple{14.00}{53.33}{8.00}& \triple{88.00}{99.33}{74.00}\\
\cmidrule{2-7}
& Llama
& \crashcell& \crashcell& \crashcell& \crashcell& \crashcell\\
\midrule

\multirow{4}{*}{30}
& Qwen
& \triple{80.94}{94.62}{71.08}& \triple{71.62}{94.60}{60.81}& \triple{69.59}{93.24}{60.14}& \triple{11.41}{40.94}{7.38}& \triple{75.68}{91.22}{68.24}\\
\cmidrule{2-7}
& Mistral
& \crashcell& \crashcell& \crashcell& \crashcell& \crashcell\\
\cmidrule{2-7}
& Phi
& \triple{91.89}{98.87}{85.14}& \triple{89.04}{98.63}{80.14}& \triple{86.90}{99.30}{77.24}& \triple{17.01}{50.34}{7.48}& \triple{91.16}{99.32}{75.51}\\
\cmidrule{2-7}
& Llama
& \crashcell& \crashcell& \crashcell& \crashcell& \crashcell\\

\bottomrule
\end{longtable}

\setlength{\tabcolsep}{3pt}
\renewcommand{\arraystretch}{1.08}

\begin{longtable}{c l c c c c c}

\caption{Average accuracy (\%, unweighted mean of the five per-task accuracies) across steering strengths, target architectures, memory defenses, and evaluators. Higher is better. Same row sets and dagger as Table~\ref{tab:full_appendix_table_syc}.}
\label{tab:full_appendix_table_util}
\\

\toprule
\textbf{$\alpha$}
& \textbf{Model}
& \textbf{No Defense}
& \textbf{Blanket Rewrite}
& \textbf{Router Gate}
& \textbf{Blanket Drop}
& \textbf{Caution}
\\
\midrule
\endfirsthead

\toprule
\textbf{$\alpha$}
& \textbf{Model}
& \textbf{No Defense}
& \textbf{Blanket Rewrite}
& \textbf{Router Gate}
& \textbf{Blanket Drop}
& \textbf{Caution}
\\
\midrule
\endhead

\multirow{4}{*}{$-6$}
& Qwen
& \triple{36.69}{39.13}{46.53}& \triple{43.20}{44.40}{52.40}& \triple{40.40}{41.20}{50.00}& \triple{12.80}{13.20}{17.60}& \triple{40.00}{42.00}{50.00} \\
\cmidrule{2-7}
& Mistral
& \crashcell& \crashcell& \crashcell& \crashcell& \crashcell\\
\cmidrule{2-7}
& Phi
& \triple{36.31}{37.07}{44.20}& \triple{43.20}{44.00}{52.00}& \triple{41.20}{40.80}{47.60}& \triple{12.00}{10.00}{17.60}& \triple{40.00}{40.80}{51.20} \\
\cmidrule{2-7}
& Llama
& \triple{23.28\dg}{19.96\dg}{22.93\dg}& \triple{29.75\dg}{25.37\dg}{32.10\dg}& \triple{25.61\dg}{22.13\dg}{29.80\dg}& \triple{6.16\dg}{3.57\dg}{7.08\dg}& \triple{15.36\dg}{12.00\dg}{18.28\dg} \\
\midrule

\multirow{4}{*}{$-2.5$}
& Qwen
& \triple{36.63}{39.05}{46.62}& \triple{42.00}{42.00}{51.60}& \triple{41.60}{40.40}{52.40}& \triple{11.20}{12.80}{18.00}& \triple{39.20}{40.80}{50.40} \\
\cmidrule{2-7}
& Mistral
& \triple{33.32}{33.35}{41.34}& \triple{41.20}{42.00}{48.00}& \triple{40.80}{41.60}{48.40}& \triple{13.60}{13.60}{19.20}& \triple{39.60}{36.40}{50.00} \\
\cmidrule{2-7}
& Phi
& \triple{35.90}{38.15}{45.05}& \triple{45.20}{47.20}{52.00}& \triple{40.40}{44.80}{50.80}& \triple{11.60}{11.60}{18.00}& \triple{34.80}{37.20}{46.80} \\
\cmidrule{2-7}
& Llama
& \triple{33.59}{34.31}{37.57}& \triple{41.00}{40.97}{47.88}& \triple{41.74}{40.93}{47.82}& \triple{11.32}{9.69}{13.76}& \triple{27.15}{28.29}{33.80} \\
\midrule

\multirow{4}{*}{$-1.5$}
& Qwen
& \triple{37.80}{39.78}{47.43}& \triple{41.60}{43.20}{52.40}& \triple{40.00}{41.60}{52.40}& \triple{12.40}{11.60}{16.80}& \triple{38.00}{39.60}{48.00} \\
\cmidrule{2-7}
& Mistral
& \triple{35.44}{34.27}{42.98}& \triple{42.00}{42.40}{50.00}& \triple{41.60}{43.60}{50.40}& \triple{16.40}{13.60}{18.80}& \triple{35.60}{39.20}{48.40} \\
\cmidrule{2-7}
& Phi
& \triple{35.94}{37.97}{44.86}& \triple{46.80}{48.80}{51.60}& \triple{38.59}{44.40}{49.60}& \triple{13.60}{12.00}{18.00}& \triple{38.00}{40.00}{50.40} \\
\cmidrule{2-7}
& Llama
& \triple{33.55}{35.35}{37.61}& \triple{44.23}{44.22}{48.68}& \triple{40.83}{43.31}{45.79}& \triple{11.34}{9.70}{14.58}& \triple{30.28}{31.08}{37.58} \\
\midrule

\multirow{4}{*}{Base (0)}
& Qwen
& \triple{38.35}{39.85}{46.78}& \triple{44.40}{44.80}{52.80}& \triple{42.80}{43.20}{52.40}& \triple{11.20}{12.40}{18.80}& \triple{40.80}{40.40}{50.00} \\
\cmidrule{2-7}
& Mistral
& \triple{33.61}{35.36}{42.99}& \triple{44.40}{42.80}{50.80}& \triple{40.80}{42.80}{49.60}& \triple{15.60}{15.60}{20.00}& \triple{40.40}{38.40}{50.40} \\
\cmidrule{2-7}
& Phi
& \triple{36.33}{38.15}{44.54}& \triple{43.60}{47.20}{54.40}& \triple{42.00}{42.40}{50.40}& \triple{12.40}{9.20}{17.20}& \triple{36.80}{39.20}{48.00} \\
\cmidrule{2-7}
& Llama
& \triple{33.97}{35.56}{38.84}& \triple{45.44}{44.22}{51.11}& \triple{41.84}{42.62}{47.52}& \triple{12.28}{10.60}{17.65}& \triple{33.37}{28.91}{41.04} \\
\midrule

\multirow{4}{*}{1.5}
& Qwen
& \triple{37.77}{39.59}{46.45}& \triple{43.60}{43.60}{50.80}& \triple{42.00}{40.40}{51.20}& \triple{13.20}{11.20}{19.20}& \triple{39.20}{40.40}{51.20} \\
\cmidrule{2-7}
& Mistral
& \triple{33.45}{34.06}{42.82}& \triple{45.20}{42.80}{46.80}& \triple{42.40}{40.40}{48.80}& \triple{13.76}{12.93}{21.42}& \triple{41.20}{36.00}{49.60} \\
\cmidrule{2-7}
& Phi
& \triple{36.45}{38.12}{46.82}& \triple{47.20}{49.60}{54.00}& \triple{42.00}{44.80}{52.00}& \triple{12.40}{10.80}{16.40}& \triple{35.60}{38.40}{48.40} \\
\cmidrule{2-7}
& Llama
& \triple{33.73}{35.33}{39.48}& \triple{38.00}{42.40}{48.00}& \triple{40.00}{42.00}{48.80}& \triple{12.26}{10.97}{14.72}& \triple{30.14}{30.51}{37.80} \\
\midrule

\multirow{4}{*}{2.5}
& Qwen
& \triple{36.25}{38.60}{46.93}& \triple{40.40}{41.20}{50.40}& \triple{40.40}{40.00}{52.00}& \triple{13.20}{12.80}{18.80}& \triple{39.60}{42.00}{50.00} \\
\cmidrule{2-7}
& Mistral
& \triple{32.85}{31.78}{42.34}& \triple{35.88}{32.58}{44.83}& \triple{35.73}{34.86}{47.12}& \triple{10.25}{12.09}{18.00}& \triple{32.80}{32.83}{45.09} \\
\cmidrule{2-7}
& Phi
& \triple{36.19}{37.82}{44.78}& \triple{46.40}{47.60}{54.80}& \triple{43.60}{44.00}{50.40}& \triple{12.00}{12.00}{16.80}& \triple{37.20}{41.60}{50.00} \\
\cmidrule{2-7}
& Llama
& \triple{33.49}{35.46}{39.74}& \triple{43.15}{42.78}{48.35}& \triple{39.03}{42.33}{45.87}& \triple{13.73}{10.51}{16.95}& \triple{25.51}{30.00}{36.48} \\
\midrule

\multirow{4}{*}{6.0}
& Qwen
& \triple{36.92}{39.14}{46.53}& \triple{42.40}{42.40}{52.40}& \triple{42.40}{41.60}{52.00}& \triple{12.00}{10.00}{19.20}& \triple{36.00}{38.80}{50.00} \\
\cmidrule{2-7}
& Mistral
& \crashcell& \crashcell& \crashcell& \crashcell& \crashcell\\
\cmidrule{2-7}
& Phi
& \triple{34.53}{37.87}{44.68}& \triple{44.00}{45.20}{49.60}& \triple{40.40}{42.00}{49.60}& \triple{11.60}{11.20}{17.20}& \triple{38.40}{39.60}{51.20} \\
\cmidrule{2-7}
& Llama
& \triple{25.00}{25.91}{34.69}& \triple{28.25\dg}{30.35\dg}{41.67\dg}& \triple{25.79\dg}{28.41\dg}{42.09\dg}& \triple{5.66}{3.76}{12.01}& \triple{17.45\dg}{19.06\dg}{28.37\dg} \\
\midrule

\multirow{4}{*}{20}
& Qwen
& \triple{32.97}{35.39}{46.33}& \triple{36.80}{38.00}{49.20}& \triple{32.00}{36.80}{50.80}& \triple{7.20}{8.00}{16.00}& \triple{31.60}{34.00}{47.20} \\
\cmidrule{2-7}
& Mistral
& \crashcell& \crashcell& \crashcell& \crashcell& \crashcell\\
\cmidrule{2-7}
& Phi
& \triple{31.01}{35.27}{43.24}& \triple{38.80}{39.20}{48.00}& \triple{35.60}{38.40}{46.00}& \triple{8.80}{8.80}{16.40}& \triple{28.80}{33.60}{43.60} \\
\cmidrule{2-7}
& Llama
& \crashcell& \crashcell& \crashcell& \crashcell& \crashcell\\
\midrule

\multirow{4}{*}{28}
& Qwen
& \triple{28.10}{29.09}{40.87}& \triple{29.45}{30.68}{41.49}& \triple{26.22}{29.44}{43.09}& \triple{6.87}{5.64}{14.96}& \triple{26.08}{25.68}{37.00} \\
\cmidrule{2-7}
& Mistral
& \crashcell& \crashcell& \crashcell& \crashcell& \crashcell\\
\cmidrule{2-7}
& Phi
& \triple{25.54}{29.91}{36.16}& \triple{31.10}{32.07}{42.94}& \triple{27.30}{29.11}{39.19}& \triple{6.00}{7.60}{13.20}& \triple{26.40}{29.20}{40.00} \\
\cmidrule{2-7}
& Llama
& \crashcell& \crashcell& \crashcell& \crashcell& \crashcell\\
\midrule

\multirow{4}{*}{30}
& Qwen
& \triple{24.65}{23.77}{36.09}& \triple{23.67}{27.68}{41.95}& \triple{22.32}{24.74}{37.81}& \triple{4.45}{5.63}{12.91}& \triple{22.15}{23.89}{37.89} \\
\cmidrule{2-7}
& Mistral
& \crashcell& \crashcell& \crashcell& \crashcell& \crashcell\\
\cmidrule{2-7}
& Phi
& \triple{24.27}{29.13}{35.18}& \triple{26.54}{31.72}{38.49}& \triple{26.26}{30.17}{38.30}& \triple{5.00}{6.67}{12.18}& \triple{23.15}{29.19}{37.49} \\
\cmidrule{2-7}
& Llama
& \crashcell& \crashcell& \crashcell& \crashcell& \crashcell\\

\bottomrule
\end{longtable}

\section{Qualitative observations and anomaly explanations}
\label{app:qualitative_observations}

During our empirical analysis, several edge-case behaviors and model-specific anomalies emerged that explain the data variance observed in Table~\ref{tab:full_appendix_table_syc}.

\subsection{Blanket Rewrite raises pooled sycophancy on Llama}
On Llama-3.1-8B the \textit{Blanket Rewrite} condition shows a higher pooled sycophancy rate than \textit{No Defense} at every coefficient, including $\alpha=0$ and the negative coefficients, so the effect is a property of the rewrite template on this model rather than of steering. One candidate explanation is that the added provenance clauses lengthen and reframe each memory in a way that Llama treats as an instruction to use it; another is that the clauses change which conflict-task items the judges read as ``used the preference''. We have not run the ablations (clause-by-clause removal, per-task decomposition with annotated examples) that would separate these, so we present the inversion as an observation, not an explained mechanism.

\subsection{Evaluator disagreement}
Qualitative inspection suggests that evaluator divergence is concentrated in ambiguous, non-committal responses. In inspected Blanket Drop examples, DeepSeek and Gemini Flash tended to assign lower sycophancy scores when the target model avoided explicit agreement, whereas GPT-4o-mini more often penalized responses that failed to directly refute the user's false premise. These observations suggest evaluator-dependent treatment of hedging rather than a consistent signal of defense failure and motivate reporting results across multiple judges.

\section{Quantifying Structural Degradation via Cap and Artifact Rates}
\label{app:structural_collapse}
\subsection{Architecture-dependent steering stability}
\label{sec:stability}
Structural degradation occurs at architecture-dependent magnitudes. We call a cell \emph{healthy} when at most 20\% of its answers are degenerate (repetition loops, empty or near-empty text; Appendix~\ref{app:collapse_analysis}), the pre-registered threshold below which a cell is claimable, and \emph{crashed} when 80\% or more are degenerate in every condition. Mistral 7B is healthy from $\alpha=-2.5$ to $2.5$ and crashed at $\alpha=-6$ and $\alpha\ge6$. Llama 3.1 8B is healthy from $-2.5$ to $2.5$, exceeds the threshold at $\alpha=-6$ in every condition and at $\alpha=+6$ in three of five (Appendix~\ref{app:collapse_analysis}; those cells are shown in the tables with a dagger and excluded from claims), and crashed at $\alpha\ge20$. Qwen 2.5 and Phi-3.5 Mini stay under the threshold across the whole ladder, although Phi's answers hit the 512-token limit in 60\% of cases at $\alpha=6$ and 87--92\% at $\alpha\ge20$, so its high-$\alpha$ cells are judged on truncated text. Steering magnitude therefore cannot be transferred across architectures without checking generation health first.

To characterize failure modes under increasingly strong activation steering, we separately track degradation of target-model outputs and artifacts introduced during evaluation. We quantify these effects using two metrics: \textit{Cap Rate} and \textit{Artifact Rate}. These metrics are analyzed separately because degradation of the target model and distortion of the LLM-as-a-Judge measurement process represent distinct failure modes.

\textit{Cap rate} is the fraction of a cell's generations that reached the 512-token \texttt{max\_new\_tokens} limit (the judge then reads a truncated answer):
\begin{equation}
    \mathrm{CapRate} = \frac{N_{\text{hit cap}}}{N_{\text{generated}}} \times 100\%.
\end{equation}
\textit{Artifact rate} is the fraction of generations that are degenerate: empty or under 20 characters (the prompts ask for an explained recommendation, so such outputs are non-answers rather than terse correct answers; the rule is a fixed, conservative filter), or flagged by a repetition test (a single token or short $n$-gram dominating the output, or low type--token diversity over a long answer):
\begin{equation}
    \mathrm{ArtifactRate} = \frac{N_{\text{degenerate}}}{N_{\text{generated}}} \times 100\%.
\end{equation}
Both are computed on the generated text before judging, per cell and per task; degenerate rows are excluded from every rate's denominator. The pre-registered rule is that a cell with artifact rate above 20\% is reported but not claimable; we additionally label a coefficient as a crash for a model when 80\% or more of its answers are degenerate in every condition. The 20\% threshold is a reporting rule fixed before grading, not a demonstrated bias bound: survivors can be a biased subset below it as well. For that reason the released tables carry the degenerate fraction of every cell next to its rate, paired tests use only items healthy in both cells, and for the headline cells (under 2\% degenerate) the conditional rate and the all-items rate (degenerate answers counted as non-sycophantic and incorrect) differ by less than one point.

At $\alpha=6$ (No Defense, 50 items per task) Qwen 2.5 has a cap rate of 4.4\% and an artifact rate of 0\%, while Llama 3.1 8B has a cap rate of 23.2\% and an artifact rate of 17.6\%: truncation and degeneration do not coincide. At $\alpha=20$ Qwen 2.5 is at 7.6\% cap and 0\% artifact, Phi-3.5 Mini at 87--92\% cap and under 2\% artifact (long but coherent answers, judged truncated), and Llama 3.1 8B and Mistral 7B are at 100\% artifact and are reported as crashes. Mistral 7B is at 85--96\% artifact at $\alpha=-6$ in all five conditions.

Extreme steering can also cause complete structural failure, characterized by repetitive, malformed, or otherwise unusable generations. Such cases are reported as model crashes rather than interpreted as successful reductions in sycophancy. Consequently, apparent decreases in sycophancy at extreme steering coefficients should not be interpreted as improved defense efficacy without first verifying generation validity.

\section{Assets, licenses, and attribution}
\label{app:assets}

All models, datasets, and evaluation endpoints used in this work are pre-existing public assets. We introduce no new models or datasets. This section lists each asset, its version identifier, its source, and its license or terms of use.

\subsection{Target models}

The four steered architectures are publicly released instruction-tuned checkpoints, obtained through the Hugging Face Hub and run locally under runtime activation steering. No weights were modified or redistributed.

\begin{table}[h]
\centering
\small
\caption{Target model checkpoints and licenses.}
\label{tab:model-licenses}
\begin{tabular}{@{}llll@{}}
\toprule
Model & Checkpoint identifier & Provider & License \\
\midrule
Llama 3.1 8B & \texttt{meta-llama/Llama-3.1-8B-Instruct}  & Meta      & Llama 3.1 Community License \\
Qwen 2.5 7B  & \texttt{Qwen/Qwen2.5-7B-Instruct}          & Alibaba   & Apache 2.0 \\
Mistral 7B   & \texttt{mistralai/Mistral-7B-Instruct-v0.3}& Mistral AI& Apache 2.0 \\
Phi-3.5 Mini & \texttt{microsoft/Phi-3.5-mini-instruct}   & Microsoft & MIT \\
\bottomrule
\end{tabular}
\end{table}

The Llama 3.1 Community License is not a standard open-source license: it imposes an acceptable-use policy and attribution requirements on derived work. Our use is non-commercial academic evaluation, we redistribute no weights or fine-tuned derivatives, and we believe this use falls within the permitted scope.

\subsection{Datasets}

\begin{itemize}
  \item \textbf{Steering-vector extraction set.} $N=100$ paired completions from \texttt{meg-tong/sycophancy-eval} on the Hugging Face Hub, the answer-region sycophancy evaluation released with Sharma et al.~\cite{sharma2023sycophancy}. Dataset card license: \textbf{MIT}.
  \item \textbf{Downstream evaluation benchmark.} All reported results are computed on the 1{,}550-item MemSyco-Bench release \cite{memsycobench2026} (\texttt{MemSyco-Bench/MemSyco-Bench} on the Hugging Face Hub), used without modification to the item set; the benchmark's \texttt{[type, status]} annotations are removed from prompts and the conversation transcript is not shown. Dataset card license: \textbf{MIT}.
  \item \textbf{Judge prompts.} The task-specific judge prompts are reproduced verbatim from the MemSyco-Bench release and hashed at load time.
\end{itemize}

\subsection{Evaluator endpoints}

Judge scoring was performed asynchronously through the OpenRouter API using the \texttt{deepseek/deepseek-v4-flash}, \texttt{openai/gpt-4o-mini}, and \texttt{google/gemini-3.7-flash} endpoints. These are accessed as hosted services under their respective providers' terms of service rather than under a redistributable license. Model outputs were used solely for automated grading of generations produced in this study; no evaluator outputs were used to train or fine-tune any model.

\subsection{Attribution}

The four checkpoints (Table~\ref{tab:model-licenses}) are Llama-3.1-8B-Instruct \cite{llama3,llama31card}, Qwen2.5-7B-Instruct \cite{qwen25,qwen25card}, Mistral-7B-Instruct-v0.3 \cite{mistral7b,mistral7b_v03}, and Phi-3.5-mini-instruct \cite{phi3,phi35mini}; the model cards document the exact checkpoints, the reports the model families. The steering method is contrastive activation addition \cite{rimsky2024caa}; the extraction data are from Sharma et al.~\cite{sharma2023sycophancy}; the evaluation benchmark and judge prompts are from Xiang et al.~\cite{memsycobench2026}. The choice of an intermediate layer is motivated by, but not taken from, Wang et al.~\cite{wang2025sycophancy}.

\section{Broader impacts}
\label{app:broader_impacts}
Positive impacts: evidence on which memory defenses preserve useful personalization while reducing agreement with stored misconceptions, and a documented warning that context ablation buys its safety numbers by discarding memory. Negative impacts and dual use: the same steering pipeline that we use as a manipulation ($\alpha>0$) can make a deployed model more sycophantic; we release vectors and code for open-weight models that are already public and add no capability beyond published contrastive activation addition, but we note the dual use explicitly. LLM-as-a-judge sycophancy rates should not be read as safety certification: three judges disagree by up to 45 points on the same answers (Blanket Drop cells, where GPT-4o-mini labels far more hedged answers sycophantic than the other two judges; Section~\ref{sec:disagreement}), and the pooled rate is dominated by one task. We did not evaluate whether memory filtering treats users' stated beliefs unevenly across groups; a gate that drops ``beliefs'' more readily than ``facts'' could suppress legitimate minority positions, and this deserves its own study.

\end{document}